\pdfoutput=1
\documentclass[11pt]{article}

\usepackage[]{EMNLP2022}

\usepackage{times}
\usepackage{latexsym}

\usepackage[T1]{fontenc}
\usepackage[utf8]{inputenc}

\usepackage{microtype}

\usepackage{color}
\usepackage{soul}
\usepackage{booktabs}
\usepackage{url}
\usepackage{amsmath}
\usepackage{danudefs}
\usepackage{adjustbox}
\usepackage{amsmath}
\usepackage{amsthm}
\usepackage{esvect}
\usepackage{multirow}
\usepackage{array}
\usepackage{subcaption}
\usepackage{braket}

\usepackage[english]{babel}
\usepackage{hyperref}
\addto\captionsenglish{%
}
\addto\extrasenglish{%
}

\usepackage{todonotes}

\makeatletter
\if@todonotes@disabled

\else

\fi
\makeatother

\usepackage{xifthen}

\DeclareMathOperator*{\minimise}{minimise}

\title{Gender Bias in Meta-Embeddings}

\author{Masahiro Kaneko$^{1}$ \quad
        Danushka Bollegala$^{2,3}$\Thanks{ Danushka Bollegala holds concurrent appointments as a Professor at University of Liverpool and as an Amazon Scholar. This paper describes work performed at the University of Liverpool and is not associated with Amazon.} \quad
        Naoaki Okazaki$^{1}$ \\
        $^1$Tokyo Institute of Technology \quad
        $^2$University of Liverpool \quad
        $^3$Amazon \\
        {\tt masahiro.kaneko@nlp.c.titech.ac.jp} \\
        {\tt danushka@liverpool.ac.uk} \quad
        {\tt okazaki@c.titech.ac.jp}
}

\begin{document}
\maketitle
\begin{abstract}
%Combining multiple source embeddings to create meta-embeddings is considered effective to obtain more accurate embeddings.
Different methods have been proposed to develop meta-embeddings from a given set of source embeddings. 
However, the source embeddings can contain unfair gender-related biases, and how these influence the meta-embeddings has not been studied yet.
We study the gender bias in meta-embeddings created under three different settings:
(1) meta-embedding multiple sources without performing any debiasing (Multi-Source No-Debiasing),
(2) meta-embedding multiple sources debiased by a single method (Multi-Source Single-Debiasing), and
(3) meta-embedding a single source debiased by different methods (Single-Source Multi-Debiasing).
Our experimental results show that meta-embedding amplifies the gender biases compared to input source embeddings.
We find that debiasing not only the sources but also their meta-embedding is needed to mitigate those biases.
Moreover, we propose a novel debiasing method based on meta-embedding learning where we use \emph{multiple} debiasing methods on a \emph{single} source embedding and then create a single unbiased meta-embedding.
\end{abstract}

\section{Introduction}

% DB: Tell that there is a trade-off between debiasing and performance. This should be connected to experiments section where we describe debiasing datasets and performance related datasets. 
% Summarise the experimental results (trends) here.

%Meta-embedding~\cite{yin-schutze-2016-learning,Bao:COLING:2018,Bollegala:IJCAI:2018} is the task of combining pre-trained source word embeddings to create more accurate and wide-coverage word embeddings.
Various pre-trained word embeddings have been successfully used as features for representing input texts in many NLP tasks~\cite{Dhillon:2015,Mnih:HLBL:NIPS:2008,Collobert:2011,Huang:ACL:2012,Milkov:2013,Pennington:EMNLP:2014}. 
Combining multiple word embeddings leads to more accurate and exhaustive meta-embeddings in terms of vocabulary, learned expressions etc~\cite{Yin:ACL:2016}.
%Different methods have been proposed such as concatenation (CONC), averaging (AVG), locally linear embedding (LLE), global linear (GLE) and autoencoders (AEME) to create meta-embeddings from given sets of source embeddings.
For example, there are meta-embedding methods that use the average of multiple embeddings~\cite{Coates:NAACL:2018}, concatenate multiple embeddings~\cite{Bollegala:met-concat}, use locally-linear~\cite{Bollegala:IJCAI:2018} or global~\cite{Yin:ACL:2016} projections, or use autoencoders~\cite{Bao:COLING:2018}.
%For example, there are meta-embedding methods that use the average of multiple embeddings ~\cite[\textbf{AVG};][]{Coates:NAACL:2018}, concatenate multiple embeddings (\textbf{CONC}), use locally-linear ~\cite[\textbf{LLE};][]{Bollegala:IJCAI:2018} or global~\cite[\textbf{GLE};][]{Yin:ACL:2016} projections, and use the intermediate layer of an autoencoder~\cite[\textbf{AEME};][]{Bao:COLING:2018}.

However, the source embeddings can contain unfair gender-related biases~\cite{barrett-etal-2019-adversarial,Xie:NIPS:2017,Elazar:EMNLP:2018,Li:2018ab}.
% and to the best of our knowledge, no prior study exists on how such biases are amplified (or not)  and how to debias during the meta-embedding process.
% Therefore, we will investigate the social biases encoded in the source embeddings and their meta-embedding.
% We will conduct an extensive systematic study covering a wide range of meta-embedding learning methods and source embeddings. 
%\section{Background}
%\label{sec:background}
To address these drawbacks, various debiasing methods have been proposed in the literature.
For example, many projection-based methods have been proposed to eliminate biases in static word embeddings~\cite{Zhao:2018ab,kaneko-bollegala-2019-gender,wang-etal-2020-double}.
\newcite{Tolga:NIPS:2016} proposed a hard-debiasing (\textbf{HARD}) method that projects gender-neutral words into a subspace, which is orthogonal to the gender dimension defined by a list of gender-definitional words.
% Adversarial learning methods~\cite{Xie:NIPS:2017,Elazar:EMNLP:2018,Li:2018ab} for debiasing first encode the inputs, and then two classifiers are jointly trained -- one predicting the target task (for which we must ensure high prediction accuracy) and the other protected attributes (that must not be easily predictable). 
% However,  \newcite{Elazar:EMNLP:2018} showed that adversarial learning alone does not guarantee invariant representations for the protected attributes.
\newcite{ravfogel-etal-2020-null} proposed iterative Null-space Projection (\textbf{INLP}) debiasing.
They found that iteratively projecting word embeddings to the null space of the gender direction improves the debiasing performance.
\newcite{Kaneko:EACL:2021a} proposed dict-debiasing (\textbf{DICT}) -- a method for removing biases from pre-trained word embeddings using dictionaries, without requiring access to the original training resources or any knowledge regarding the word embedding algorithms used.
%Adversarial learning methods~\cite{Xie:NIPS:2017,Elazar:EMNLP:2018,Li:2018ab}.
%However,  \newcite{Elazar:EMNLP:2018} showed that adversarial learning alone does not guarantee invariant representations for the protected attributes.

\begin{figure*}[t]
  \centering
  \includegraphics[width=0.9\textwidth]{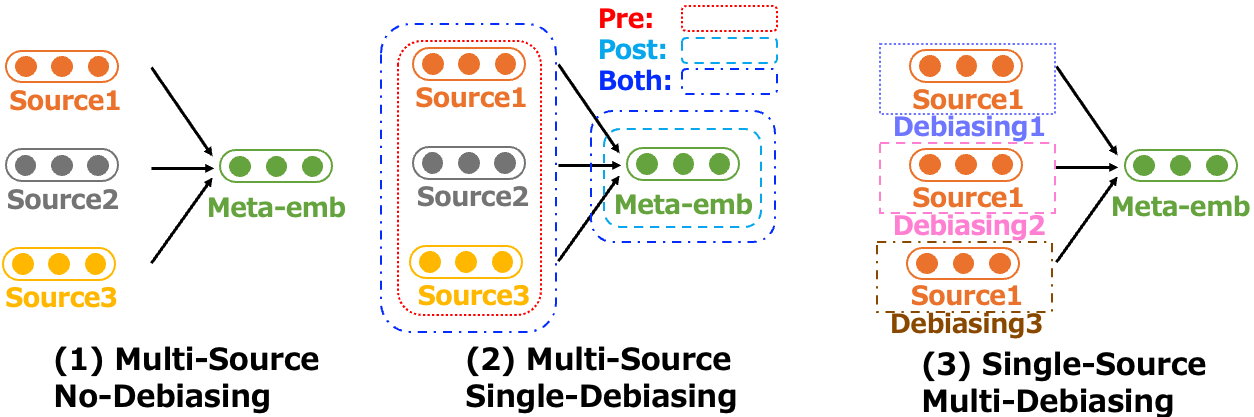}
  \caption{We investigate the bias in three types of meta-embeddings: Multi-Source No-Debiasing, Multi-Source Single-Debiasing and Single-Source Multi-Debiasing. Src denotes the source embeddings and the boxes represent target of debiasing. pre, post, and both indicate as to what stage the debiasing is performed.%Multi-Source No-Debiasing learns meta-embedding from multiple kinds of source embeddings using meta-embedding learning method. Multi-Source Single-Debiasing learns meta-embedding from multiple kinds of source embeddings. Multi-Source Single-Debiasing learns meta-embedding by adapting a single debiasing method to multiple types of source embeddings. There are three types of adaptation: adaptation to the source embeddings, adaptation to the learned meta-embeddings, and adaptation of the debiasing method to both. Multi-Source Single-Debiasing adapts several different debiasing methods from a single source embedding and learns meta-embeddings from them.
  }
  \label{fig:all_methods}
\end{figure*}

On the other hand, to the best of our knowledge, the effect on gender bias due to meta-embedding that uses multiple sources has not been investigated.
Even if we had perfectly debiased the individual source embeddings, some meta-embedding methods such as averaging do \emph{not} guarantee debiased meta-embeddings as we prove in \autoref{sec:math}.
In this study, we classify meta-embeddings into the following three types from the viewpoint of debiasing and analyze them: 
(1) \textbf{Multi-Source No-Debiasing}: meta-embeddings created from multiple source embeddings without any debiasing;
(2)  \textbf{Multi-Source Single-Debiasing}: meta-embeddings created from multiple source embeddings debiased by a single debiasing method;
(3)  \textbf{Single-Source Multi-Debiasing}: meta-embedding from the same source embedding, debiased using different debiasing methods.\footnote{It is also possible to adapt multiple debiasing methods to multiple source embeddings and learn meta-embedding from them, but this is not the focus of this study because it would increase the vector size (total dimensionality of source embeddings $\times$ number of debiasing methods) and computation cost tremendously. For example, in this case, there are four 300-dimensional word embeddings and three debiasing methods, so (4 $\times$ 300) $\times$ 3 results in a 3600-dimensional vector.}
Multi-Source Single-Debiasings were examined by debiasing: each source embeddings (\textbf{pre}), the learned meta-embeddings (\textbf{post}), and both source embeddings and meta-embeddings (\textbf{both}).
\autoref{fig:all_methods} shows how a meta-embedding is learned for those methods.
%for Multi-Source No-Debiasing, Multi-Source Single-Debiasing and Single-Source Multi-Debiasing, respectively.

These methods are agnostic to types of meta-embedding learning algorithms and demonstrate different aspects of debiasing effects.
%The investigation purposes for Multi-Source No-Debiasing, Multi-Source Single-Debiasing, and Single-Source Multi-Debiasing are different as follows.
We use Multi-Source No-Debiasings to investigate bias in meta-embeddings learned using existing approaches.
The purpose of Multi-Source Single-Debiasing is to investigate how to effectively debias existing meta-embeddings.
In Single-Source Multi-Debiasing, we combine the same embeddings debiased by different methods to investigate whether debiasing methods can complement each other's strengths and weaknesses to obtain more effective debiased embeddings.
To the best of our knowledge, no studies have been proposed that combine multiple debiasing methods.
%With respect to multi-debiasing, in the existing debiasing research, it is not clear whether the combination of multiple debiasing methods can achieve better debiased embeddings compared to using a single debiasing method by complementing their strengths and weaknesses.
%In addition, we propose a \textit{multi-debiasing} method that combines various debiasing methods with meta-embedding learning methods to complement their strengths and weaknesses and achieve better debiased embeddings compared to using a single debiasing method.

%We use three debiasing methods (hard-debiasing, INLP and dict-debiasing) and five meta-embeddings (CONC, AVG, GLE, LLE, and AEME) in our study.
We use three debiasing methods and five meta-embeddings in our study.
We focus on gender bias, since there are several methods~\cite{Tolga:NIPS:2016,ravfogel-etal-2020-null,Kaneko:EACL:2021a} that can be used to combine debiasing methods and datasets~\cite{Caliskan2017SemanticsDA,Zhao:2018ab,du-etal-2019-exploring} that can be examined in different ways.
Experimental results show that the gender bias is amplified by meta-embedding methods without any treatment for debiasing.
Moreover, the gender bias increases with the number of source embeddings used in the meta-embedding.

Interestingly, we can successfully debias meta-embeddings without losing their superiority of the performance improvements in two out of three word embedding benchmarks.
The Multi-Source Single-Debiasing results indicate that debiasing both source embeddings and meta-embeddings is the best practice in two out of three bias evaluation benchmarks.
We also demonstrate that Single-Source Multi-Debiasing performs better than using only one debiasing method in all three bias evaluation benchmarks.
It can be seen as a debiasing method that uses an ensemble of existing debiasing methods via a meta-embedding framework to create more reliable unbiased embeddings than if we had used a single debiasing method.
This is an important result given that there exists a broad range of debiasing methods proposed in the NLP community based on different principles and complementary strengths, yet no single best method exist~\cite{meade-etal-2022-empirical,czarnowska-etal-2021-quantifying}.

\section{Meta-Embedding Learning Methods}
\label{sec:ME-methods}

Depending on whether debiasing methods are applied on source embeddings or their meta-embedding, three variants can be identified: Multi-Source No-Debiasing, Multi-Source Single-Debiasing and Single-Source Multi-Debiasing.
To explain these settings further, let us consider a set of $N$ source word embeddings $s_{1}, s_{2}, \ldots, s_{N}$ respectively covering vocabularies (i.e. sets of words) $\cV_1, \cV_2, \ldots, \cV_N$.
The embedding of a word $w$ in $s_{j}$ is denoted by $\vec{s}_{j}(w) \in \R^{d_{j}}$, where $d_{j}$ is the dimensionality of $\vec{s}_{j}$.
%We can represent $s_{j}$ by an embedding matrix $\mat{E}_{j} \in \R^{d_{j} \times |\cV_j|}$.
In Multi-Source No-Debiasing, meta-embedding $\vec{m}^{\rm MSND}(w)$ for word $w$ is computed from $\vec{s}_{1}(w), \vec{s}_{2}(w), \ldots, \vec{s}_{N}(w)$ using some meta-embedding learning method, where MSND represents Multi-Source No-Debiasing.
Multi-Source Single-Debiasing and Single-Source Multi-Debiasing are obtained by applying debiasing methods described in \autoref{sec:debiasing-methods}.
Let us denote debiasing for word embedding $\vec{s}_{j}(w)$ as $\vec{d}(\vec{s}_{j}(w))$.
In Multi-Source Single-Debiasing, there are three types of debiasing possibilities: \emph{pre}, \emph{post}, and \emph{both}.
In \emph{pre}, the debiased source embeddings $\vec{d}(\vec{s}_{1}(w)), \vec{d}(\vec{s}_{2}(w)), \ldots, \vec{d}(\vec{s}_{N}(w))$ are used to computed $\vec{m}^{\rm MSSDpre}(w)$, where MSSD represents Multi-Source Single-Debiasing.
In \emph{post}, debiasing is performed on the learned meta-embeddings as in $\vec{m}^{\rm MSSDpost}(w) = \vec{d}(\vec{m}^{\rm MSND}(w))$.
In \emph{both}, debiasing is performed for both pre and post, as in $\vec{m}^{\rm MSSDboth}(w) = \vec{d}(\vec{m}^{\rm MSSDpre}(w))$.
In Single-Source Multi-Debiasing, we use different debiasing methods for the same source embedding as in $\vec{d}_1(\vec{s}_{j}(w)), \vec{d}_2(\vec{s}_{j}(w)), \ldots, \vec{d}_M(\vec{s}_{j}(w))$ to learn meta-embedding $\vec{m}^{\rm SSMD}(w)$.
Here, $M$ is the number of debiasing methods and SSMD represents Single-Source Multi-Debiasing.

The source word embeddings, in general, do not have to cover the same set of words.
Much prior work in meta-embedding learning assume a common vocabulary over all source embeddings for simplicity.
If a particular word is not covered by a source embedding, it can be assigned a zero vector, a randomly initialised vector or we could learn a regression model to predict the missing source embeddings~\cite{yin-schutze-2016-learning}.
Without loss of generality, we will assume that all words for evaluation are covered by a meta-embedding vocabulary $\cV$, which is composed by each source embedding's vocabulary $\mathcal{V}_j$, after applying any one of the above-mentioned methods.
Here $j$ represents the j-th source embedding's vocabulary. 
Each word $w$ is assumed to be included in at least one of the vocabularies $\mathcal{V}_j$, and zero embeddings are assigned for $w \notin \cV_j$.

%Next, we describe the five meta-embedding learning methods, CONC, AVG, GLE, LLE, and AEME, used to learn $\vec{m}(w)$ for Multi-Source No-Debiasing, Multi-Source Single-Debiasing and Single-Source Multi-Debiasing.

We consider five previously proposed meta-embedding learning methods for static word embeddings in this study to learn $\vec{m}(w)$ for Multi-Source No-Debiasing, Multi-Source Single-Debiasing and Single-Source Multi-Debiasing as follows:
concatenation~\cite[\textbf{CONC}][]{Bollegala:met-concat},
averaging~\cite[\textbf{AVG}][]{Coates:NAACL:2018},
globally-linear meta-embedding~\cite[\textbf{GLE}][]{yin-schutze-2016-learning},
locally-linear meta-embedding~\cite[\textbf{LLE}][]{Bollegala:IJCAI:2018}, and
averaged autoencoded meta-embeddings~\cite[\textbf{AEME}][]{Bao:COLING:2018}.
According to~\newcite{Bollegala:IJCAIb:2022}, these are the most widely-used meta-embedding learning methods.
They methods are described in detail in Appendix \S 1.

\section{Debiasing Methods for Static Word Embeddings}
\label{sec:debiasing-methods}

Different methods have been proposed in prior work for debiasing static word embeddings.
We consider the following three popular debiasing methods in this study: 
(1) hard-debiasing~\cite[\textbf{HARD}][]{Tolga:NIPS:2016}, 
(2) Iterative Null Space Projection~\cite[\textbf{INLP}][]{ravfogel-etal-2020-null}, 
and (3) dictionary-based debiasing~\cite[\textbf{DICT}][]{Kaneko:EACL:2021a}.
%\DB{I have used HARD and DICT as method names. We need to use the same names (uppercase) in the Tables later on to be consistent.}
Due to space constraints, we describe those methods in detail in Appendix \S 2.
By adapting these debiasing methods to the meta-embeddings described in \autoref{sec:ME-methods}, we investigate Multi-Source Single-Debiasing and Single-Source Multi-Debiasing.
%\DB{If we have space lets briefly explain hard-debiasing, INLP and dict-debiasing in a single paragraph here, without using equations. We can use the Supplementary to explain the methods in detail.}

%Adversarial learning methods~\cite{Xie:NIPS:2017,Elazar:EMNLP:2018,Li:2018ab} for debiasing first encode the inputs. Then two classifiers are jointly trained -- one predicting the target task (for which we must ensure high prediction accuracy) and the other protected attributes (that must not be easily predictable). 
% However,  \newcite{Elazar:EMNLP:2018} showed that adversarial learning alone does not guarantee invariant representations for the protected attributes.

\section{AVG does \emph{not} Protect HARD Debiasing}
\label{sec:math}

In general, it is difficult to mathematically analyze the gender bias in debiased source embeddings when they are meta-embedded using a particular meta-embedding learning method.
However, such an analysis is possible in the special case of the HARD debiasing for CONC and AVG, and shows that even if all source embeddings are debiased their meta-embedding might not always remain debiased.

Let us consider applying HARD to debias two sources $s_1, s_2$ independently and create their meta-embeddings separately using CONC and AVG. 
To simplify the discussion, let us assume that both $s_1$ and $s_2$ to be $k$-dimensional and having bias vector sets respectively $\{\vec{b}^{(1)}_1, \ldots, \vec{b}^{(1)}_k\}$ and  $\{\vec{b}^{(2)}_1, \ldots, \vec{b}^{(2)}_k\}$.
We debias the source embeddings of a word $w$ using HARD and obtain $\vec{d}_1(w)$ and $\vec{d}_2(w)$, given respectively by \eqref{eq:d1} and \eqref{eq:d2}.
\begin{align}
    \label{eq:d1}
    \vec{d}_1(w) = \frac{\vec{s}_1(w) - \vec{w}_{1,\cB}}{\norm{\vec{s}_1(w) - \vec{w}_{1,\cB}}}
\end{align}
\begin{align}
    \label{eq:d2}
    \vec{d}_2(w) = \frac{\vec{s}_2(w) - \vec{w}_{2,\cB}}{\norm{\vec{s}_2(w) - \vec{w}_{2,\cB}}}
\end{align}
Here, $\vec{w}_{1,\cB}$ and $\vec{w}_{2,\cB}$ denote the projected source embeddings of $w$ onto the gender subspaces in each source embedding spaces $s_1$ and $s_2$.
Let us denote the concatenated meta-embedding of $\vec{d}_1(w)$ and $\vec{d}_2(w)$ by $\vec{m}_{\rm conc}(w)$.
Consider the bias vector $\vec{b}^{(1)}_j \oplus \vec{b}^{(2)}_j$.
Because the inner-product decomposes over the individual components under vector concatenation we can simplify
$\braket{\vec{m}_{\rm conc}(w), \vec{b}^{(1)}_j \oplus \vec{b}^{(2)}_j}$ as follows:
\begin{align}
\label{eq:b1}
    \braket{\vec{d}_1(w), \vec{b}^{(1)}_j} + \braket{\vec{d}_2(w),  \vec{b}^{(2)}_j}
\end{align}
Each term in \eqref{eq:b1} are separately zero because the debiased embeddings are orthogonal to the bias vectors by construction in each source.
Therefore, concatenated meta-embedding preserves the debiasing result under HARD debiasing.

However, this is not true for other meta-embedding methods such as averaging.
To see this, consider $\braket{\vec{d}_1(w) + \vec{d}_2(w), \vec{b}^{(1)}_j + \vec{b}^{(2)}_j}$, which results in four terms as in \eqref{eq:b2}.
\begin{align}
\label{eq:b2}
    &\braket{\vec{d}_1(w), \vec{b}^{(1)}_j} +  \braket{\vec{d}_2(w), \vec{b}^{(2)}_j} \nonumber \\
    &+  \braket{\vec{d}_1(w), \vec{b}^{(2)}_j} +  \braket{\vec{d}_2(w), \vec{b}^{(1)}_j}
\end{align}
Note that the first two terms in \eqref{eq:b2} are zero because they are in the same vector space and the inner-products are taken w.r.t. to the corresponding bias vectors.
However, the last two terms in \eqref{eq:b2} are \emph{not} generally zero.
Therefore, AVG does not generally preserve the HARD debiasing result.

\section{Experiments}
\label{sec:exp}

Our goal in this paper is to evaluate whether gender bias is amplified and to what degree by the different meta-embedding learning methods.
However, this bias amplification must be considered relative to the accuracy of the semantic representations produced by those meta-embedding learning methods.
For example, a meta-embedding learning method can produce perfectly unbiased representations by mapping all words to a constant vector, which is useless for any downstream task requiring semantic representations of words.
For this reason, we conduct our evaluations using two types of datasets to evaluate gender biases in the debiased embeddings, while preserving useful semantic information necessary for downstream tasks: (a) bias evaluation datasets covering different types of gender biases (described in \autoref{sec:datasets}) and (b) word embedding benchmarks related to semantic similarity prediction and POS tagging tasks (described in \autoref{sec:embedding-datasets}).

\subsection{Bias Evaluation Benchmarks}
\label{sec:datasets}

We use Word Embedding Association Test~\cite[{\bf WEAT};][]{Caliskan2017SemanticsDA}, Word Association Test~\cite[{\bf WAT};][]{du-etal-2019-exploring} and SemBias~\cite[{\bf SB};][]{Zhao:2018ab} for bias evaluation. 
The closer the scores of all of these evaluations to 0, the less bias there is.

\paragraph{WEAT:} WEAT~\cite{Caliskan2017SemanticsDA} quantifies various biases (e.g., gender, race, and age) using semantic similarities between word embeddings.
It compares two same size sets of \emph{target} words $\cX$ and $\cY$ (e.g. European and African names), with two sets of \emph{attribute} words $\cA$ and $\cB$ (e.g. \emph{pleasant} vs. \emph{unpleasant}). 
The bias score, $s(\cX,\cY,\cA,\cB)$, for each target is calculated as follows:
\begin{align}
s(\cX,\cY,\cA,\cB) &= \sum_{\vec{x} \in \cX} k(\vec{x}, \cA, \cB) \nonumber \\
                   &- \sum_{\vec{y} \in \cY} k(\vec{y}, \cA, \cB) \\
k(\vec{t}, \cA, \cB) &= \textrm{mean}_{\vec{a} \in \cA} f(\vec{t}, \vec{a}) \nonumber \\
                     &- \textrm{mean}_{\vec{b} \in \cB} f(\vec{t}, \vec{b})
\end{align}
Here, $f$ is the cosine similarity between the word embeddings.
The one-sided $p$-value for the permutation test regarding $\cX$ and $\cY$ is calculated as the probability of $s(\cX_i,\cY_i,\cA,\cB) > s(\cX,\cY,\cA,\cB)$.
The effect size is calculated as the normalised measure given by \eqref{eq:effect}.
\begin{align}
\label{eq:effect}
\frac{\textrm{mean}_{x \in \cX} s(x, \cA,\cB) - \textrm{mean}_{y \in \cY} s(y, \cA, \cB)}{\textrm{sd}_{t \in \cX \cup \cY} s(t, \cA, \cB)}
\end{align}
WEAT can evaluate eight types of bias, and we report the average  absolute effect sizes for T4, T5 and T6, which are related to gender bias.

\paragraph{WAT:} WAT\footnote{\url{https://github.com/Yupei-Du/bias-in-wat}} is a method to measure gender bias over a large set of words~\cite{du-etal-2019-exploring}.
It calculates the gender information vector for each word in a word association graph created with Small World of Words project~\cite[SWOWEN;][]{Deyne2019TheW} by propagating information related to masculine and feminine gender pair set (e.g. \textit{she} and \textit{he}) $(w_m^i, w_f^i) \in \cL$, using a random walk approach~\cite{Zhou2003LearningWL}.
The gender information is represented as a 2-dimensional vector ($b_m$, $b_f$), where $b_m$ and $b_f$ denote respectively the masculine and feminine orientations of a word.
The gender information vectors of masculine words, feminine words, and other words are initialised respectively with vectors (1, 0), (0, 1), and (0, 0).
The bias score of a word is defined as $\log(b_m / b_f)$.
We evaluate the gender bias of word embeddings using the Pearson correlation coefficient between the bias score of each word and 
the score given by \eqref{eq:bias-score} computed as the averaged difference of cosine similarities between masculine and feminine words.
\begin{align}
\label{eq:bias-score}
\frac{1}{|\cL|} \sum_{i=1}^{|\cL|} \left( f(w, w_m^i) - f(w, w_f^i) \right)
\end{align}

\paragraph{SB:} The SB dataset~\cite{Zhao:2018ab}\footnote{\url{https://github.com/uclanlp/gn_glove}} contains three categories of word-pairs: 
(1) \textbf{Definition}, a gender-definition word pair (e.g. hero -- heroine), 
(2) \textbf{Stereotype}, a gender-stereotype word pair (e.g., manager -- secretary) 
and (3) \textbf{None}, two other word-pairs with similar meanings unrelated to gender (e.g., jazz -- blues, pencil -- pen). 
The SB metric uses the cosine similarity between the $\vv{he} - \vv{she}$ gender directional vector and $\vec{a} - \vec{b}$ for each word pair (a -- b) in the above categories to measure gender bias.
SB contains 20 Stereotype word pairs and 22 Definition word pairs and uses the Cartesian product to generate 440 instances.
\newcite{Zhao:2018ab} used a subset of 40 instances associated with two seed word-pairs, not used in the word list for training, to evaluate the generalisability of a debiasing method.
We expect high similarity scores in the Definition category and low similarity scores in the Stereotype and None categories for unbiased word embeddings.
This paper reports the percentage of times that pairs of Stereotype and None categories had the highest similarity in the subset and this score is expected to be low.

\begin{table}[t]
\begin{adjustbox}{width=\columnwidth,center}
\centering
\begin{tabular}{lccc|ccc}
\toprule
  & WEAT & WAT & SB & SL & MEN & POS \\
\midrule
W2V & 1.31 & 0.47 & 17.0 & 44.2 & 78.2 & 87.8 \\
GV  & 1.17 & \textbf{0.58} & 17.0 & 40.8 & 80.5 & 90.9 \\
FTC & 1.31 & 0.53 & 13.2 & \textbf{47.1} & 81.5 & 88.7 \\
FTW & 1.08 & 0.50 & 15.2 & 44.1 & 80.1 & 80.1 \\
%\hdashline
ALL & 1.22 & 0.52 & 15.6 & 44.1 & 80.1 & 88.8 \\
\midrule
AVG  & \textbf{1.46} & 0.53 & 18.0 & 41.7 & 80.5 & 89.7  \\
CONC  & 1.33 & \textbf{0.58} & 16.6 & 42.7 & 81.3 & \textbf{91.2} \\
LLE  & 1.39 & 0.56 &\textbf{ 30.0} & 44.2 & 80.8 & 89.0 \\
GLE & 1.31 & 0.52 & 16.8 & 43.7 & \textbf{82.1} & 87.7 \\
AEME & 1.28 & 0.53 & 11.1 & 43.7 & 81.1 & 89.1 \\
\bottomrule
\end{tabular}
\end{adjustbox}
\caption{The results of bias evaluation and word embedding benchmarks for source embeddings (W2V, GV, FTC and FTW) and meta-embeddings of Multi-Source No-Debiasing (AVG, CONC, LLE, GLE and AEME). ALL is the score to compare with the score of Multi-Source No-Debiasing, which is arithmetic mean over the score of source embeddings. \textbf{Bold} indicates the results with the highest bias and performance.}
\label{tbl:bias_in_meta}
\end{table}

\begin{table}[h]
\begin{adjustbox}{width=\columnwidth,center}
\centering
\begin{tabular}{lcccc|ccc}
\toprule
 & num & WEAT & WAT & SB & SL & MEN & POS \\
\midrule
ALL & 1 & 1.22 & 0.52 & 15.6 & 44.1 & 80.1 & 88.8 \\
\midrule
\multirow{3}{*}{AVG} &
  2 & 1.32 & 0.51 & 17.3 & 43.0 & 79.9 & 89.1 \\
& 3 & 1.38 & \textbf{0.53} & 17.5 & 42.8 & 80.0 & 89.6 \\
& 4 & \textbf{1.46} & \textbf{0.53} & \textbf{18.0} & 41.7 & \textbf{80.5} & \textbf{89.7} \\
\midrule
\multirow{3}{*}{CONC} &
  2 & 1.26 & 0.53 & 16.4 & 43.1 & 80.3 & 90.3 \\
& 3 & 1.30 & 0.57 & 16.5 & 42.2 & 80.6 & 90.8 \\
& 4 & \textbf{1.33} & \textbf{0.58} & \textbf{16.6} & 42.7 & \textbf{81.3} & \textbf{91.2} \\
\midrule
\multirow{3}{*}{LLE} &
  2 & 1.25 & 0.52 & 21.7 & 43.3 & 80.3 & 87.9 \\
& 3 & 1.29 & 0.54 & 25.2 & 43.9 & 80.5 & 88.4 \\
& 4 & \textbf{1.39} &\textbf{ 0.56} &\textbf{30.0} & \textbf{44.2} & \textbf{80.8} & \textbf{89.0} \\
\midrule
\multirow{3}{*}{GLE} &
  2 & 1.26 & 0.50 & 16.0 & 43.3 & 81.0 & 87.6 \\
& 3 & 1.29 & 0.50 & 16.3 & 43.5 & 81.3 & 87.6 \\
& 4 & \textbf{1.31} & \textbf{0.52} & \textbf{16.8} & 43.7 & \textbf{82.1} & 87.7 \\
\midrule
\multirow{3}{*}{AEME} &
  2 & 1.24 & 0.52 & 12.0 & 43.3 & 80.5 & 88.9 \\
& 3 & 1.25 & 0.52 & 12.6 & 43.5 & 80.8 & 89.0 \\
& 4 & \textbf{1.28} & \textbf{0.53} & 11.1 & 43.7 & \textbf{81.1} & \textbf{89.1} \\
\bottomrule
\end{tabular}
\end{adjustbox}
\caption{The results of bias evaluation and word embedding of benchmarks of Multi-Source No-Debiasing with AVG, CONC, LLE, GLE and AEME using different number of source embeddings. This is the result of arithmetic mean scores for each number. Here, num=1 represents the arithmetic mean of the results of all source embeddings. \textbf{Bold} indicates the results with the highest bias and performance in num=2, 3, 4, considering num=1, 2, 3, 4.}
\label{tbl:source_num}
%\vspace{-3mm}
\end{table}

\begin{table}[t]
\begin{adjustbox}{width=\columnwidth,center}
\centering
\begin{tabular}{llccc|ccc}
\toprule
\multicolumn{2}{c}{Method} & WEAT & WAT & SB & SL & MEN & POS \\
\midrule
\multicolumn{2}{c}{ALL} & 1.22 & 0.52 & 15.6 & 44.1 & 80.1 & 88.8 \\
\midrule
\multicolumn{2}{c}{HARD} & 0.93 & 0.45 &  \underline{7.7} & 44.2 & 80.0 & 88.5 \\
\multicolumn{2}{c}{INLP} & \underline{0.91} & \underline{0.43} & 12.2 & 43.6 & 79.2 & 88.7 \\
\multicolumn{2}{c}{DICT} & 0.97 & 0.51 & 12.9 & \textbf{47.2} & \textbf{82.1} & \textbf{88.8} \\
\midrule
\multirow{3}{*}{HARD} &
  pre  & 0.93 & 0.48 & 9.7 & 43.2 & 79.5 & 87.1 \\
& post & \underline{0.84} & 0.50 & 9.2 & 42.4 & \textbf{79.9} & \textbf{87.9} \\
& both & 0.86 & \underline{0.40} & \underline{9.1} & \textbf{44.1} & 79.7 & 87.7 \\
\midrule
\multirow{3}{*}{INLP} &
  pre  & 0.95 & 0.49 & 13.2 & 42.2 & \textbf{79.3} & 88.2 \\
& post & 0.91 & 0.48 & 12.5	& 41.1 & \textbf{79.3} & \textbf{88.7} \\
& both & \underline{0.86} & \underline{0.37} & \underline{12.2}	& \textbf{44.2} & 79.1 & 88.6 \\
\midrule
\multirow{3}{*}{DICT} &
  pre  & 1.01 & 0.50 & 14.2 & 46.1 & \textbf{84.9} & 89.0 \\
& post & 0.97 & 0.51 & 13.8 & 45.3 & 85.2 & \textbf{90.1} \\
& both & \underline{0.89} & \underline{0.44} & \underline{14.1} & \textbf{46.3} & 84.6 & 89.9 \\
\bottomrule
\end{tabular}
\end{adjustbox}
\caption{The results of bias evaluation and word embeddings benchmarks of source embeddings (ALL), debiased source embeddings (HARD, INLP, DICT) and meta-embeddings of Multi-Source Single-Debiasing (pre, post, both). Here, pre indicates debiasing source embeddings then learning meta-embeddings, post indicates debiasing the meta-embeddings, and both indicates debiasing both source and meta-embeddings. pre, post and both average the scores using the five meta-embedding methods. ALL, HARD, INLP and DICT scores are the average results of W2V, GV, FTC and FTW. \underline{Underline} represents the most debiased results and \textbf{Bold} shows the highest performance.}
\label{tbl:debiased_meta}
%\vspace{-3mm}
\end{table}

\subsection{Word Embedding Benchmarks}
\label{sec:embedding-datasets}

We use SimLex~\cite[{\bf SL};][]{J15-4004}, {\bf MEN}~\cite{P12-1015} and CoNLL-2003 POS tagging~\cite[{\bf POS tagging};][]{Zhao:2018ab} as word embedding benchmarks. 
The higher these scores, the better the performance.

\paragraph{Semantic Similarity:} 
The semantic similarity between two words is calculated as the cosine similarity between their word embeddings and compared against the human ratings using the Spearman correlation coefficient.
Following prior work, we use SL~\cite{J15-4004} and MEN~\cite{P12-1015} for evaluations.
%The following datasets are used: Word Similarity 353~\cite[{\bf WS};][]{Finkelstein:2001}, {\bf SimLex}~\cite{J15-4004}, Rubenstein-Goodenough~\cite[{\bf RG};][]{Rubenstein:1965}, {\bf MTurk}~\cite{Halawi:2012}, rare words~\cite[{\bf RW};][]{W13-3512} and {\bf MEN}~\cite{P12-1015}.

\paragraph{POS tagging:}
To evaluate the performance in a downstream task that uses word/meta embeddings as input representations, we evaluate the performance of POS tagging of the model initialised by the pre-trained word embedding.
We use the CoNLL-2003 dataset~\cite{tjong-kim-sang-de-meulder-2003-introduction} for training and evaluating the POS tagger, implemented as a single LSTM with a 100-dimensional hidden layer.
All the weights and biases of LSTM are initialized from $\mathcal{U}(-\sqrt{k}, \sqrt{k})$ where $\mathcal{U}$ is a uniform distribution and $k = \frac{1}{\rm hidden\_size}$.
We optimise the model using SGD with a learning rate of 0.1.
We set the batch size to 32 and report results of the model on the test data. The model with the best performance was selected using the development data in 10 epochs.
We use WEAT T4 as the development data.

% \paragraph{Word Analogy:} 
% In word analogy, we predict $d$ that completes the proportional analogy ``$a$ is to $b$ as $c$ is to what?'', for four words $a$, $b$, $c$ and $d$.
%  We use CosAdd~\cite{Levy:CoNLL:2014}, which determines $d$ by maximising the cosine similarity between the two vectors ($\vec{b} - \vec{a} + \vec{c}$) and $\vec{d}$.
%  Following \newcite{Zhao:2018ab}, we evaluate on {\bf MSR}~\cite{mikolov-yih-zweig:2013:NAACL-HLT} and {\bf Google} analogy datasets~\cite{NIPS2013_5021} as shown in \autoref{tbl:sem}.

\subsection{Settings}

In our experiments, we use the following publicly available pre-trained word embeddings as the source embeddings:
Word2Vec\footnote{\url{https://code.google.com/archive/p/word2vec/}}~\cite[\textbf{W2V};][]{NIPS2013_5021} is 300-dimensional embeddings for 3M words trained on Google News corpus,
GloVe\footnote{\url{https://github.com/stanfordnlp/GloVe}}~\cite[\textbf{GV};][]{Glove} is 300-dimensional embeddings for 2.2M words trained on the Common Crawl and FastText\footnote{\url{https://fasttext.cc/docs/en/english-vectors.html}}~\cite[\textbf{FTC} and \textbf{FCW};][]{bojanowski2017enriching} are 300-dimensional embeddings for 2M words trained on Common Crawl and Wikipedia.

We used the publicly available code by the original authors for \textbf{HARD}\footnote{\url{https://github.com/tolga-b/debiaswe}}, \textbf{INLP}\footnote{\url{https://github.com/shauli-ravfogel/nullspace_projection}} and \textbf{DICT}\footnote{\url{https://github.com/kanekomasahiro/dict-debias}} debiasing methods with the default hyperparameters and word lists for training used in the original implementations.
Debiasing requires less than half an hour in all experiments on a GeForce RTX 2080 Ti GPU.

\subsection{Gender Bias in Meta-Embeddings}
\label{sec:exp:bias}

% \begin{table*}[t]
% %\begin{adjustbox}{width=\textwidth,center}
% \centering
% \small
% \begin{tabular}{lccc|ccc}
% \toprule
%  & WEAT & WAT & SB & SL & MEN & POS \\
% \midrule
% num=1 & 1.10 $\pm$ 0.13 & 0.52 $\pm$ 0.04 & \textbf{15.6} $\pm$ 1.81 & \textbf{44.1} $\pm$ 2.57 & 80.1 $\pm$ 1.39 & 88.8 $\pm$ 1.44 \\
% num=2 & 1.08 $\pm$ 0.20 & 0.50 $\pm$ 0.20 & 18.1 $\pm$ 3.13 & 42.5 $\pm$ 1.23 & 80.1 $\pm$ 2.04 & 88.2 $\pm$ 2.32 \\
% num=3 & \textbf{1.07} $\pm$ 0.15 & 0.51 $\pm$ 0.14 & 19.2 $\pm$ 5.93 & 42.6 $\pm$ 0.91 & 80.3 $\pm$ 1.83 & 88.4 $\pm$ 3.23 \\
% num=4 & 1.10 $\pm$ 0.19 & \textbf{0.49} $\pm$ 0.10 & 18.6 $\pm$ 6.81  & 43.0 $\pm$ 0.93 & \textbf{80.5} $\pm$ 2.00 & \textbf{88.9} $\pm$ 1.98 \\
% \bottomrule
% \end{tabular}
% %\end{adjustbox}
% \caption{The result of meta-embeddings with different number of source embeddings. This is the result of averaging the scores for each number. Here, num = 1 represents the average of the results of all source embeddings.}
% \label{tbl:source_num}
% %\vspace{-3mm}
% \end{table*}

To study how different Multi-Source No-Debiasing methods amplify the gender bias in the source embeddings, in
\autoref{tbl:bias_in_meta} we compare source and meta-embeddings using the datasets described in \autoref{sec:exp}.
From \autoref{tbl:bias_in_meta} we see that different source embeddings express different levels of gender biases.
Among the source embeddings, we see that GV has the highest bias in WEAT, WAT and SB.
On the other hand, FTC has the best performance in SL and MEN, whereas GV has the best performance in POS.
Here, \textbf{ALL} is the arithmetic mean of the scores for the four source embeddings W2V, GV, FTC and FTW, which simulates the setting where sources are used separately without creating any meta-embeddings.
We use ALL to compare with the results of Multi-Source No-Debiasing.

Among the Multi-Source No-Debiasing methods, we see that LLE has a lower bias on WEAT and WAT, but its performance is considerably worse in MEN and POS.
Ideally, we would prefer Multi-Source No-Debiasing methods that combine the information in multiple sources, while not aggregating or amplifying any gender bias present in the source embeddings.
In this regard, we can see that AEME, which uses autoencoders to learn meta-embeddings, has a lower bias as well as better performance.
This result aligns well with prior proposals where \newcite{kaneko-bollegala-2019-gender} used autoencoders to debias static word embeddings.
Moreover, it has been shown that autoencoding improves pre-trained word embeddings by making them more isotropic~\cite{kaneko-bollegala-2020-autoencoding}, which might explain the superior performance of AEME as a meta-embedding learning method.

Considering that Multi-Source No-Debiasing methods use multiple source embeddings as the input, an interesting open question is whether more sources result in more biased meta-embeddings.
To study this relationship empirically, in \autoref{tbl:source_num} we use varying numbers of source embeddings and create meta-embeddings.
%In this study, we use CONC because it has both high bias and high performance, making it easier to analyse trends.
For example, the row corresponding to num=2 in \autoref{tbl:source_num} shows the setting where we use two out of the available four source embeddings to create meta-embeddings. 
This results in six ($_4C_2$) different meta-embeddings produced in num=2 setting.
%Because we have five different meta-embedding learning methods, we end up with 30 ($5 \times 6$) meta-embeddings in num=2 setting. 
We evaluate each of those meta-embeddings for the bias and semantic representation ability and report the arithmetic mean.
Note that num=1 setting corresponds to the previously described \textbf{ALL} baseline, which reports the average scores when each source embedding is evaluated individually, without creating their meta-embeddings.
num=4 are the same as AVG, CONC, LLE, GLE, AEME in \autoref{tbl:bias_in_meta}.

From \autoref{tbl:source_num} we see that, except for AEME in SB, the gender bias is amplified when more sources are used in the meta-embedding process.
Moreover, the average performances of meta-embeddings increase with the number of sources.
In Multi-Source No-Debiasing, we can see that increasing the number of source embeddings amplifies the bias as well as improving the task performance.
%However, recall that some meta-embedding learning methods perform worse than others.
%Moreover, due to negative transfer, the best performance by a meta-embedding learning method can correspond to not using all but a subset of the sources.
%For example, as shown in the supplementary materials, AEME reports the best performance among the different meta-embedding learning methods and obtains better performance than num=1 in all SL, MEN, POS tasks with different subsets of source embeddings.

\subsection{Debiasing vs. Meta-Embedding}
\label{sec:debias-vs-meta}

Next, we study the effectiveness of Multi-Source Single-Debiasing using different debiasing methods described in \autoref{sec:debiasing-methods} for removing unfair gender bias from Multi-Source No-Debiasing.
Note that debiasing and meta-embedding learning methods can be applied to a given set of source embeddings in an arbitrary order.
Here we consider three settings: \textbf{pre} (first debias the sources and then create their meta-embedding),
\textbf{post} (first create meta-embedding of the sources and then debias it), and
\textbf{both} (apply debiasing to the source as well as meta-embeddings).
For \textbf{pre}, \textbf{post} and \textbf{both} of Multi-Source Single-Debiasing, we use five meta-embedding learning methods and reported their arithmetic mean scores.\footnote{If the results for each meta-embedding method are listed in \autoref{tbl:debiased_meta}, there will be 5 rows for each of pre, post and both, and 9 (number of pre, post and both) $\times$ 5 (number of meta-embedding) + 5 (rows other than pre, post and both) will make a huge table of 50 rows, so for the reason of space, the meta-embedding scores are averaged.}
\autoref{tbl:debiased_meta} shows the arithmetic mean results of applying HARD, INLP, and DICT-debiasing compared to all the source embeddings and the results of Multi-Source Single-Debiasing.
%First, it can be seen that applying debiasing methods to source embeddings results in debiasing compared to source embeddings.
First, we see that all debiasing methods reduce the gender bias in the source embeddings compared to ALL.
%By applying DICT-debiasing to SL and MEN, we can see that the performance is significantly improved.
In particular, DICT-debiasing not only debiases the source embeddings but also improves their performance, as can be seen from the evaluations on SL and MEN datasets.

%The debiasing results for meta-embeddings show that all the results are better than the average results for source embeddings.
%This shows that the existing debiasing methods are effective for embeddings made from multiple embeddings.
%In particular, both debiasings are effective, indicating that it is important to perform debiasing on both source embeddings and meta-embeddings learned from them.
%Furthermore, we can see that debiasing on both embeddings does not degrade SL, MEN, and POS performance.

We see that HARD, INLP and DICT debiasing methods for a single source are still effective at debiasing meta-embeddings created from multiple sources.
Furthermore, \textbf{both} outperforms \textbf{pre} and \textbf{post} for debiasing meta-embeddings except HARD both on WEAT, while not degrading performance on SL, MEN and POS.
Although we do not show the individual results of meta-embedding in \autoref{tbl:debiased_meta}, out of 45 results (i.e. 3 debiasing methods $\times$ 5 meta-embeddings $\times$ 3 test data) there are 5 cases where \textbf{both} under-performs in bias evaluation to \textbf{pre} or \textbf{post}. 
Moreover, the tendency for \textbf{both} to obtain the best debiasing results does not depend on the meta-embedding method used.

\begin{figure}[t]
  \centering
  \includegraphics[width=\columnwidth]{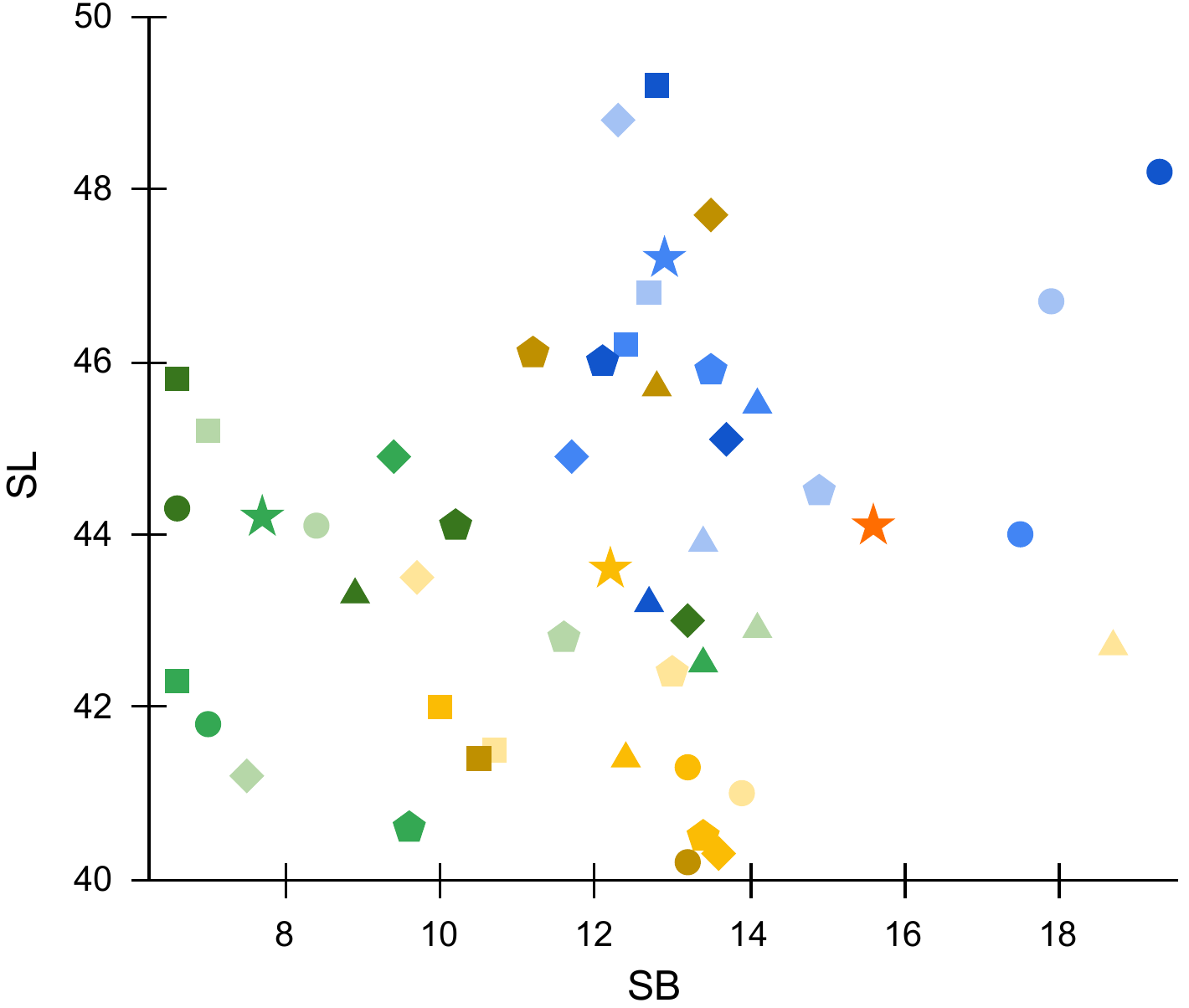}
  \caption{Scores for SB and SL. SB score is horizontal axis and SL score is vertical axis.}
  \label{fig:plot}
\end{figure}

Because scores in \autoref{tbl:debiased_meta} are averaged over meta-embedding methods, to further analyse effects due to each meta-embedding method, we plot all combinations in \autoref{fig:plot}, where $x$-axis shows SB scores (lower values indicate debiased embeddings) and $y$-axis shows SL scores (higher values indicate accurate embeddings).
The shapes of the points denote the meta-embedding methods: source - star; AVG - circle; CONC - square; LLE - triangle; GLE - pentagon; and AEME - diamond, respectively.
The colors indicate the debiasing methods: No-Debiasing - orange; HARD - green; INLP - yellow; DICT - blue, respectively.
The density of the colours indicates pre: light, post: intermediate, and both: dark, respectively.

Except for some results of DICT, meta-embeddings are debiased regardless of the debiasing method used compared to their source embeddings.
In most cases, CONC results in the most debiased meta-embeddings compared to the debiased source embeddings.
In addition, HARD and DICT of CONC show improved SL performance compared to their source embeddings.
The performance of CONC is considered to be higher than that of source embeddings because the number of dimensions of CONC is larger and more expressive than that of source embeddings.

\begin{table}[t]
\begin{adjustbox}{width=\columnwidth,center}
\centering
\begin{tabular}{lcccc|ccc}
\toprule
 & Emb. & WEAT & WAT & SB & SL & MEN & POS \\
\midrule
%Single & 0.98 & 0.46 & 10.9 & 45.0 & 80.4 & 88.7 \\
%\midrule
% AVG & 0.85 & 0.45 & 9.7 & 44.8 & 79.6 & 88.3 \\
% CONC & 0.88 & 0.42 & 9.4 & 45.0 & 80.7 & \textbf{89.6} \\
% LLE & \textbf{0.58} & \textbf{0.29} & 10.4 & 44.6 & 80.6 & 86.7 \\
% GLE & 0.90 & 0.40 & 11.8 & 43.3 & \textbf{81.6} & 87.2 \\
% AEME & 0.95 & 0.38 & \textbf{8.8} & \textbf{45.1} & 80.5 & 89.0 \\
\multirow{4}{*}{\rotatebox[origin=c]{90}{Source}}
& W2V & 1.31 & 0.47 & 17.0 & 44.2 & 78.2 & 87.8 \\
& GV  & 1.17 & 0.58 & 17.0 & \textbf{40.8} & 80.5 & 90.9 \\
& FTC & 1.31 & 0.53 & 13.2 & \textbf{47.1} & 81.5 & 88.7 \\
& FTW & 1.08 & 0.50 & 15.2 & \textbf{44.1} & 80.1 & 80.1 \\
\midrule
\multirow{4}{*}{\rotatebox[origin=c]{90}{Debiased}}
& W2V & 1.08 & 0.46 & 11.7 & 44.2 & \textbf{80.8} & \textbf{90.8} \\
& GV  & 1.01 & 0.52 & 11.9 & 40.5 & \textbf{81.5} & 90.7 \\
& FTC & 1.14 & 0.51 & 12.5 & 46.8 & 78.9 & 89.4 \\
& FTW & 0.94 & 0.47 & 13.4 & 44.0 & 82.0 & \textbf{81.5} \\
\midrule
\multirow{4}{*}{\rotatebox[origin=c]{90}{SSMD}}
& W2V & \underline{0.98} & \underline{0.44} & \underline{9.8} & \textbf{44.3} & 80.7 & 90.5 \\
& GV  & \underline{0.90} & \underline{0.32} & \underline{11.0} & 40.5 & \textbf{81.5} & \textbf{91.1} \\
& FTC & \underline{1.05} & \underline{0.41} & \underline{8.9} & 45.3 & \textbf{81.8} & \textbf{89.8} \\
& FTW & \underline{0.78} & \underline{0.44} & \underline{12.1} & 43.0 & \textbf{82.3} & 81.0 \\
\bottomrule
\end{tabular}
\end{adjustbox}
\caption{The results of bias evaluation and word embedding benchmarks of source, debiased source and Single-Source Multi-Debiasing embeddings (SSMD). The results of each meta-embedding learning method are arithmetic means in SSMD. \underline{Underline} represents the most debiased results per embedding method.}
\label{tbl:meta_debias}
%\vspace{-3mm}
\end{table}

\subsection{The Same Source Embeddings, Different Debiasing}
\label{sec:multi-debias}

In \autoref{sec:debias-vs-meta} we observed that each debiasing method has its own strengths and weaknesses in removing bias-related information and preserving useful semantic information in word embeddings.
Motivated by this, we propose Single-Source Multi-Debiasing -- given a pre-trained source embedding, we first apply different debiasing methods to create multiple debiased versions of that source embedding and subsequently meta-embed them.
Specifically, we create debiased versions of a source embedding using HARD, INLP, and DICT debiasing methods separately, 
and then use AVG, CONC, LLE, GLE to create corresponding meta-embeddings.
%As an example, \autoref{fig:meta_debias} shows the multi-debiasing of W2V.

\autoref{tbl:meta_debias} shows the results for the original source embeddings (Source), debiased source embeddings (Debiased source) and Single-Source Multi-Debiasing embeddings.
%Therein, multi-debias is the average of the results obtained by applying one debiasing method to one source embedding (4 source embeddings x 3 debiasing methods).
Here, Single-Source Multi-Debiasing shows the arithmetic mean of the scores of the five meta-embedding learning methods.\footnote{If the results each meta-embedding method are listed in \autoref{tbl:meta_debias}, there will be 5 rows for each word embedding, and 4 $\times$ 5 + 9 will make a huge table of 29 rows, thus due to limited space the scores are averaged.}
Moreover, debiased source shows the arithmetic mean of the scores of adapting the three debiasing methods separately to compare the methods using multiple debiasing methods.
We see that the bias evaluation of all Single-Source Multi-Debiasing is better than source and debiased source.
This indicates that Single-Source Multi-Debiasing improves the overall debiasing performance by taking into account the strengths and weaknesses of each debiasing method.
Furthermore, the highest number of scores in SL, MEN and POS indicates that Single-Source Multi-Debiasing is able to learn the highest quality embeddings.
Although we do not put the individual results of meta-embedding in \autoref{tbl:meta_debias}, out of 60 results (i.e. 4 source embeddings $\times$ 5 meta-embeddings $\times$ 3 test data), Single-Source Multi-Debiasing underperforms only in 4 cases compared to the debiased sources.

\section{Conclusion}

We studied the gender bias due to meta-embedding under three settings: (1) Multi-Source No-Debiasing, (2) Multi-Source Single-Debiasing (3) Single-Source Multi-Debiasing created from static word embeddings as sources.
Our experimental results show that although meta-embedding of Multi-Source No-Debiasing improves performance over the input source embeddings, at the same time, it amplifies the unfair gender bias encoded in the source embeddings.
Furthermore, the level of gender bias encoded in a meta-embedding increases with the number of source embeddings used.
We found that Multi-Source Single-Debiasing using previously proposed debiasing methods for static word embeddings can be effectively used to debiase meta-embeddings as well.
Furthermore, we proposed Single-Source Multi-Debiasing that combines the outputs from multiple debiasing methods and then create a single embedding via a meta-embedding learning method.

%\section*{Acknowledgements}
%This paper is based on results obtained from a project, JPNP18002, commissioned by the New Energy and Industrial Technology Development Organization (NEDO).
% Entries for the entire Anthology, followed by custom entries

\section{Limitations}
\label{sec:limitations}

In this paper, we limited our investigation to meta-embedding learning methods applicable to static word embeddings because they are still extensively used in various NLP applications for input representation, particularly in resource/energy constrained devices without GPUs due to their relatively lightweight nature compared to contextualised embeddings obtained from large-scale neural language models~\cite{strubell-etal-2019-energy}.
However, there has been recent work studying the gender bias in contextualised embeddings~\cite{Zhao:2019a,Vig:2019,bordia-bowman-2019-identifying,may-etal-2019-measuring,Kaneko:EACL:2021b,kaneko2021unmasking,Kaneko:MLM:2022,Zhou:2022,schick2020self}.
On the other hand, learning meta-embeddings of contextualised embeddings is relatively underdeveloped~\cite{poerner-etal-2020-sentence}.
Therefore, we defer the study of gender bias in contextualised meta-embeddings to future work.
Furthermore, in future, we plan to study other types of social biases such as racial and religious biases in meta-embeddings.

\section{Ethical Considerations}
\label{sec:ethics}

The goal of our paper was to study the gender bias in various meta-embeddings created in three different settings.
We did not manually annotate novel social bias datasets, proposed novel bias evaluation measures nor debiasing methods.
Therefore, we do not see any ethical issues arising due to data annotation, or via proposals of novel evaluation metrics or debiasing methods.

The gender biases we considered in this paper cover only binary gender.
The bias evaluation in word embeddings used in our paper can evaluate only binary gender.
However, gender biases have been reported related to non-binary gender as well~\cite{cao-daume-iii-2020-toward,dev-etal-2021-harms}.
Studying the non-binary gender for debiasing meta-embeddings is an essential next step.

This paper does not cover all debiasing methods for word embeddings~\cite{kaneko-bollegala-2019-gender,wang-etal-2020-double} and does not guarantee results with any given debiasing method.
Furthermore, it should be noted that there may be bias when using debiased meta-embeddings in a downsteream task.
It is known that the results of task-independent bias evaluation do not necessarily coincide with the bias evaluation in the downstream task~\cite{goldfarb-tarrant-etal-2021-intrinsic,Cao:2022}.

\bibliographystyle{acl_natbib}
\bibliography{custom}

\clearpage
\appendix

\section{Meta-Embedding Learning Methods}
\label{sec:sup:ME-methods}

\subsection{Concatenation (CONC)}
\label{sec:conc}

One of the simplest approaches to create a meta-embedding under the unsupervised setting is vector concatenation~\cite{Bao:COLING:2018,Yin:ACL:2016,Bollegala:IJCAI:2018}.
Denoting concatenation by $\oplus$, we can express the concatenated meta-embedding, $\vec{m}_{\mathrm{conc}}(w) \in \R^{d_1 + \ldots + d_N}$, of a word $w \in \cV$ by \eqref{eq:conc}.
\begin{align}
\label{eq:conc}
\vec{m}_{\mathrm{conc}}(w) &= \vec{s}_{1}(w)\oplus  \ldots \oplus \vec{s}_{N}(w) \nonumber \\
& = \oplus_{j=1}^{N} \vec{s}_{j}(w)
\end{align}
\newcite{AAAI:2016:Goikoetxea} showed the concatenation of word embeddings learnt separately from a corpus and the WordNet produces superior word embeddings.
%Moreover, applying Principal Component Analysis on the concatenated embeddings further improved the performance on word similarity tasks.
However, one disadvantage of using concatenation to produce meta-embeddings is that it increases the dimensionality of the meta-embedding space, which is the sum of the dimensionalities of the sources.
%\cite{Yin:ACL:2016} post-processed the meta-embeddings created by concatenating the source embeddings using SVD to reduce the dimensionality.
%However, applying SVD often results in degradation of accuracy in the meta-embeddings compared to the original concatenated version~\newcite{Bollegala:IJCAI:2018}.

% It is easier to see that concatenation does not remove any information that is already covered by the source embeddings.
% However, it is not obvious under what conditions concatenation could produce a meta-embedding that is superior to the input source embeddings.
% A more recent work by \cite{Bollegala:NAACL:2021} shows that concatenation minimises the pairwise inner-product \cite[(PIP)][]{Yin:2018} loss between the source embeddings and an idealised meta-embedding.
% PIP loss has been shown to be directly related to the dimensionality of a word embedding, and has been used as a criterion for selecting the optimal dimensionality for static word embeddings.
% They propose a weighted variant of concatenation where the dimensions of each source are linearly weighted prior to concatenation.
% The weight parameters can be learnt in an unsupervised manner by minimising the empirical PIP loss.

\subsection{Averaging (AVG)}
\label{sec:avg}

Source embeddings are trained independently and can have different dimensionalities.
Even when the dimensionalities do agree, vectors that lie in different vector spaces cannot be readily averaged.
However, rather surprisingly, \newcite{Coates:NAACL:2018} showed that accurate meta-embeddings can be produced by first zero-padding source embeddings as necessary to bring them to the same dimensionality and then by averaging them to create $\vec{m}_{\mathrm{avg}}(w)$ as given by \eqref{eq:avg} when some orthogonality conditions are satisfied by the embedding spaces.
\begin{align}
\label{eq:avg}
\vec{m}_{\mathrm{avg}}(w) = \frac{1}{N} \sum_{j=1}^{N} \vec{s}^{*}_j(w)
\end{align}
Here, $\vec{s}^{*}_j(w)$ is the zero-padded version of $\vec{s}_j(w)$ such that its dimensionality is equal to $\max(d_1, \ldots, d_N)$.
In contrast to concatenation, averaging has the desirable property that the dimensionality of the meta-embedding is upper-bounded by $\max(d_1, \ldots, d_N) < \sum_{j=1}^{N} d_j$.

\subsection{Linear Projections (GLE and LLE)}
\label{sec:proj}

In their pioneering work on meta-embedding, \newcite{Yin:ACL:2016} proposed to project source embeddings to a common space via source-specific linear transformations, which they refer to as 1\texttt{TO}N.
They require that the meta-embedding of a word $w$, $\vec{m}_{\text{1\texttt{TO}N}}(w) \in \R^{d_{m}}$, to be reconstructed from each source embedding, $\vec{s}_j(w)$ of $w$.
For that they use a linear projection matrix, $\mat{A}_j \in \R^{d_j \times d_m}$, from $s_j$ to the meta-embedding space as given by \eqref{eq:1ton}.
\begin{align}
\label{eq:1ton}
\hat{\vec{s}}_j(w) = \mat{A}_j \vec{m}_{\text{1\texttt{TO}N}}(w)
\end{align}
Here, $\hat{\vec{s}}_j(w)$ is the reconstructed source embedding of $w$ from the meta-embedding $\vec{m}_{\text{1\texttt{TO}N}}(w)$.
Next, the squared Euclidean distance between the source- and meta-embeddings is minimised over all words in the intersection of the source vocabularies, subjected to Frobenius norm regularisation as in \eqref{eq:loss-1ton}.

\begin{align}
\footnotesize
\label{eq:loss-1ton}
\minimise_{\substack{\forall_{j=1}^{N} \mat{A}_j \\  \forall_{w \in \cV}\ \vec{m}_{\text{1\texttt{TO}N}}(w)}} \sum_{j=1}^{N} \alpha_j \left( \sum_{w \in \cV} \norm{\hat{\vec{s}}_j(w)  - \vec{s}_j(w)}_2^2 + \norm{\mat{A}_j}_{F}^{2} \right)
\end{align}
They use different weighting coefficients $\alpha_j$ to account for the differences in accuracies of the sources.
They determine $\alpha_j$ using the Pearson correlation coefficients computed between the human similarity ratings and cosine similarity computed using the each source embedding between word pairs on the \newcite{MCdataset} dataset.
The parameters can be learnt using stochastic gradient descent, alternating between projection matrices and meta-embeddings.

\newcite{muromagi-etal-2017-linear} showed that by requiring the projection matrices to be orthogonal (corresponding to the Orthogonal Procrustes Problem), the accuracy of the learnt meta-embeddings is further improved.
However,  1\texttt{TO}N requires all words to be represented in all sources.
%To overcome this limitation, they predict the source embedding for missing words as described in \autoref{sec:definition}.

Assuming that a single \emph{global linear} (\textbf{GLE}) projection can be learnt between the meta-embedding space and each source embedding as done by \newcite{Yin:ACL:2016} having all words in all sources is a more vital requirement. 
\newcite{Bollegala:IJCAI:2018} relaxed this requirement by learning \emph{locally linear} (\textbf{LLE}) meta-embeddings.
To explain this method further let us consider computing the LLE-based meta-embedding, $\vec{m}_{\mathrm{LLE}}(w)$, of a word $w \in \cV_1 \cap \cV_2$ using two sources $s_1$ and $s_2$.
First, they compute the set of nearest neighbours, $\cN_j(w)$, of $w$ in $s_j$ and represent $w$ as the linearly-weighted combination of its neighbours by a matrix $\mat{A}$ by minimising \eqref{eq:LLE-1}.
\par\nobreak
\begin{align}
\label{eq:LLE-1}
\footnotesize
 \minimise_{\mat{A}} \sum_{j=1}^{2} \sum_{w \in \cV_1 \cap \cV_2} \norm{\vec{s}_j(w) - \sum_{w' \in \cN_j(w)} A_{ww'} \vec{s}_j(w)}_2^2
\end{align}
They use AdaGrad to find the optimal $\mat{A}$.
Next, meta-embeddings are learnt by minimising \eqref{eq:LLE-2} using the learnt neighbourhood reconstruction weights in $\mat{A}$ are  preserved in a vector space common to all source embeddings. 
\par\nobreak
\begin{align}
\label{eq:LLE-2}
\footnotesize
\sum_{w \in \cV_1 \cap \cV_2} \norm{\vec{m}_{\mathrm{LLE}}(w) - \sum_{j=1}^{2}\sum_{w' \in \cN_j(w)} C_{ww'}\vec{m}_{\mathrm{LLE}}(w')}_2^2
\end{align}
Here, $C_{ww'} = A_{ww'} \sum_{j=1}^{2} \mathbb{I}[w' \in \cN_j(w)]$, where $\mathbb{I}$ is the indicator function which returns 1 if the statement evaluated is True.
Optimal meta-embeddings can then be found by solving an eigendecomposition of the matrix $(\mat{I} - \mat{C})\T(\mat{I} - \mat{C})$, where $\mat{C}$ is the matrix formed by arranging $C_{ww'}$ as the $(w,w')$ element.
This approach has the advantage of not requiring all words to be represented by all sources, thereby obviating the need to predict missing source embeddings prior to meta-embedding.

\subsection{Autoencoding (AEME)}
\label{sec:ae}

\newcite{Bao:COLING:2018} modelled meta-embedding learning as an \emph{autoencoding} problem where information embedded in different sources is integrated at different levels to propose Averaged Autoencoded meta-embedding (\textbf{AEME}).

Consider two sources $s_1$ and $s_2$, which are encoded respectively by two encoders $E_1$ and $E_2$.
AEME of $w$ is computed as the $\ell_2$ normalised average of the encoded source embeddings as in \eqref{eq:AEME}.
\par\nobreak
\begin{align}
\label{eq:AEME}
\vec{m}_{\mathrm{AEME}}(w) = \frac{E_1(\vec{s}_1(w)) + E_2(\vec{s}_2(w))}{\norm{E_1(\vec{s}_1(w)) + E_2(\vec{s}_2(w))}_2}
\end{align}
Two independent decoders, $D_1$ and $D_2$, are trained to reconstruct the two sources from the meta-embedding.
$E_1, E_2, D_1$ and $D_2$ are jointly learnt to minimise the weighted reconstruction loss given by \eqref{eq:AE-loss}.
\par\nobreak
{\footnotesize
\begin{align}
\label{eq:AE-loss}
\minimise_{E_1,E_2,D_1,D_2} \sum_{w \in \cV_1 \cap \cV_2} \Large( & \lambda_1 \norm{\vec{s}_1(w) - D_1(E_1(\vec{s}_1(w)))}_2^2 + \nonumber \\
								& \lambda_2 \norm{\vec{s}_2(w) - D_2(E_2(\vec{s}_2(w)))}_2^2 \Large)
\end{align}
}
The weighting coefficients $\lambda_1$ and $\lambda_2$ can be used to assign different emphases to reconstruct the two sources and are tuned using a validation dataset.
In comparison to methods that learn globally or locally linear transformations~\cite{Bollegala:IJCAI:2018,Yin:ACL:2016}, autoencoders learn nonlinear transformations.

AEME can only use two source embeddings to learn a meta-embedding.
Therefore, in cases with more than two source embeddings, we adapt  AEME to learn meta-embeddings from meta-embeddings created from two source embeddings and other source embeddings.
%Their proposed autoencoder variants outperform 1\texttt{TO}N and 1\texttt{TO}N+ on multiple benchmark tasks.

\section{Debiasing Methods}
\label{sec:sup:debiasing-methods}

\subsection{Hard-debiasing}

\newcite{Tolga:NIPS:2016}~proposed a post-processing approach that projects gender-neutral words to a subspace, which is orthogonal to the gender direction defined by a list of gender-definitional words to reduce the gender stereotypes embedded inside pre-trained word representations.
Their hard-debiasing method computes the gender direction as the vector difference between the embeddings of the corresponding gender-definitional words.
They denote the $n$-dimensional pre-trained word embedding of a word $w$ by $\vec{w} \in \R^n$.
$W$ is a set of pre-trained word embeddings $\vec{w}_1, \vec{w}_2, \dots \vec{w}_v$.
Here, $v$ is a vocabulary size.
% They refer to words associated with gender (e.g., \emph{she}, \emph{actor}) as gender-definitional words, and the remainder gender-neutral.
% They proposed a \emph{hard-debiasing} method where the gender direction is computed as the vector difference between the embeddings of the corresponding gender-definitional words, and a \emph{soft-debiasing} method, which balances the objective of preserving the inner-products between the original word embeddings, while projecting the word embeddings into a subspace orthogonal to the gender definitional words. 
% They use a seed set of gender-definitional words to train a support vector machine classifier, and use it to expand the initial set of gender-definitional words.
% Both hard and soft debiasing methods ignore gender-definitional words during the subsequent debiasing process, and focus only on words that are \emph{not} predicted as gender-definitional by the classifier. 
% Therefore, if the classifier erroneously predicts a stereotypical word as a gender-definitional word, it would not get debiased.
To reduce the gender bias, it is assumed that the $n$-dimensional basis vectors in the $\mathbb{R}^{n}$ vector space spanned by the pre-trained word embeddings to be $\vec{b}_{1}, \vec{b}_{2}, \ldots, \vec{b}_{n}$.
Moreover, without loss of generality, the subspace spanned by the subset of the first $k (<n)$ basis vectors $\vec{b}_{1}, \vec{b}_{2}, \ldots, \vec{b}_{k}$  denoted to be $\cB \subseteq \R^{n}$.
The projection $\vec{v}_{\cB}$ of a vector $\vec{v} \in \R^{n}$ onto $\cB$ can be expressed using the basis vectors as in \eqref{eq:vb}.
%\par\nobreak
%{\small
%\vspace{-5mm}
\begin{align}
\label{eq:vb}
\vec{v}_{\cB} = \sum_{j=1}^{k} (\vec{v}\T\vec{b}_{j}) \vec{b}_{j}
\end{align}
%}
To show that $\vec{v} - \vec{v}_{\cB}$ is orthogonal to $\vec{v}_{\cB}$ for any $\vec{v} \in \cB$, $\vec{v} - \vec{v}_{\cB}$ is expressed using the basis vectors as given in \eqref{eq:v_vb}.
%\par\nobreak
%{\small
%\vspace{-5mm}
\begin{align}
\vec{v} - \vec{v}_{\cB} &=  \sum_{i=1}^{n} (\vec{v}\T\vec{b}_{i}) \vec{b}_{i} - \sum_{j=1}^{k} (\vec{v}\T\vec{b}_{j}) \vec{b}_{j} \nonumber \\
                        &= \sum_{i=k+1}^{n} (\vec{v}\T\vec{b}_{i}) \vec{b}_{i}
\label{eq:v_vb}
\end{align}
%}
It can be seen that there are no basis vectors in common between the summations in \eqref{eq:vb} and \eqref{eq:v_vb}.
Therefore, $\vec{v}_{\cB} \T (\vec{v} - \vec{v}_{\cB}) = 0$ for $\forall \vec{v} \in \cB$.

To identify gender subspace, a set of $n$ a masculine and feminine word pairs $D_1, D_2, ..., D_n \subset W^2$ is defined, here each pair of words indicates gender.
The average vector $\mu$ of the defining sets is represented in \eqref{eq:mu}.
\begin{align}
\label{eq:mu}
\mu_i := \sum_{\vec{w} \in D_i} \frac{\vec{w}}{|D_i|}
\end{align}
Let the bias subspace $\cB$ be the first $k$ rows of Singular Value Decomposition SVD ($\bf C$),
\begin{align}
\label{eq:phi}
\textbf{C} := \sum_{i=1}^n \sum_{\vec{w} \in D_i} (\vec{w} - \mu_i)^{\top}(\vec{w} - \mu_i) / |D_i|
\end{align}
Hard-debiasing removes the bias by zero projection of all neutral words into the bias subspace $\cB$.
Then, debiased embedding $\vec{d}_{\rm hard}(\vec{w})$ is represented in \eqref{eq:hard}:
\begin{align}
\label{eq:hard}
\vec{d}_{\rm hard}(\vec{w}) := \frac{\vec{w} - \vec{w}_{\cB}}{\norm{\vec{w} - \vec{w}_{\cB}}}
\end{align}
% \begin{align}
% \label{eq:hard_debias}
% \vec{w} := \frac{\vec{w} - \vec{w}_B}{\norm{\vec{w} - \vec{w}_B}}
% \end{align}

\citet{gonen-goldberg-2019-lipstick} showed that hard-debiasing does not completely remove gender biases from embeddings.
On the other hand, the motivation of Single-Source Multi-Debiasing is to complement weaknesses of each debiasing method by combining the various debiasing methods.
Therefore, we use hard-debiasing, a method known to produce incomplete debiasing, to investigate whether we could overcome its limitations by meta-embedding with source embeddings produced by other debiasing methods.

\subsection{Iterative Null-space Projection (INLP)}

INLP was proposed by \newcite{ravfogel-etal-2020-null} to remove bias in pre-trained embeddings by iteratively projecting onto null-space.
They train multiple linear classifiers to detect bias in a pre-trained embedding and remove information by projecting the embedding into the null space of the weights of each linear classifier.
By adapting multiple classifiers, it is possible to remove bias by projecting embeddings into the null space using dozens of directions based on the data.

First, let $C$ be the parameter of the linear classifier that detects the bias in the word embeddings.
For example, in gender bias detection, this linear classifier will classify whether the representation is feminine or masculine.
The fact that the classifier cannot classify the embedding as feminine or masculine, as in \eqref{eq:c}, means that there is no bias in the embedding.
\begin{align}
\label{eq:c}
C(P_{N(C)} \vec{w}) = 0  \quad \forall \vec{w}
\end{align}
Here, $W$ is projected to a space orthogonal to C, i.e., null-space, so that the decision boundary of $C$ cannot detect the bias.
The null-space at $C$ is defined as $N(C) = \lbrace \vec{w} | C \vec{w} = 0 \rbrace$, and the projection matrix onto $N(C)$ is $P_{N(C)}$.

Relations in a multidimensional space can be captured in multiple linear directions (hyperplanes).
Therefore, it is not sufficient to project the embedding into the null space of a single linear classifier.
To solve this problem, they adapt the classifier iteratively.
The projection matrix $P$ is iterated $m$ times as $P = P_{N(C_{m})} P_{N(C_{m-1})} \ldots P_{N(C_{1})}$.
$C_i$ is learned from $W_{i-1}$ projected into the null space of $C_{i-1}$.
Debiased embedding $\vec{d}_{\rm inlp}(\vec{w})$ is represented in \eqref{eq:inlp}:
\begin{align}
\label{eq:inlp}
\vec{d}_{\rm inlp}(\vec{w}) := P\vec{w}
\end{align}

\subsection{Dict-debiasing}

\newcite{Kaneko:EACL:2021a} proposed dict-debiasing --  a method for debiasing pre-trained word embeddings using dictionary definitions.
This method does not need the types of biases to be pre-defined in the form of word lists and learns the constraints that must be satisfied by unbiased word embeddings automatically from dictionary definitions of the words.

This method assumes that a dictionary $\cD$ containing the definition, $g(w)$ of $w$, is given. 
If the pre-trained embeddings distinguish among the different senses of $w$, then the gloss for the corresponding sense of $w$ in the dictionary can be used as $g(w)$.
%However, the majority of word embedding learning methods do not produce sense-specific word embeddings. 
%In this case, we can either use all glosses for $w$ in $\cD$ by concatenating or select the gloss for the dominant (most frequent) sense of $w$\footnote{Prior work on debiasing static word embeddings do not use contextual information that is required for determining word senses. Therefore, for comparability reasons, we do neither.}.
%Without any loss of generality, in the remainder of this paper, we will use $s(w)$ to collectively denote a gloss selected by any one of the criteria mentioned above with or without considering the word senses.
%Next, we define the objective functions optimised by the dict-debiasing method to learn unbiased word embeddings.
Given $\vec{w}$, which is the word embedding of a word $w$, they model the debiasing process as the task of learning an encoder $E(\vec{w}; \vec{\theta}_e)$ that returns an $m (\leq n)$-dimensional debiased version of $\vec{w}$. 
To preserve the dimensionality of the input embeddings, they set $m = n$.

To preserve semantic information during the debiasing process, they decode the encoded version of $\vec{w}$ using a decoder $D_{c}$ parametrised by $\vec{\theta}_{c}$ and define $J_{c}$ to be the reconstruction loss given by \eqref{eq:Jc}.
\begin{align}
 J_{c}(w) = \norm{\vec{w} - D_{c}(E(\vec{w}; \vec{\theta}_{e}); \vec{\theta}_{c})}_{2}^{2}
 \label{eq:Jc}
\end{align}

To ensure that the encoded version of $\vec{w}$ is similar to $g(w)$, $g(w)$ is represented by a sentence embedding vector $\vec{g}(w) \in \R^{n}$.
For the simplicity, they use the smoothed inverse frequency~\cite[SIF;][]{Arora:ICLR:2017} for creating $\vec{g}(w)$.
SIF computes the embedding of a sentence as the weighted average of the pre-trained word embeddings of the words in the sentence, where the weights are computed as the inverse unigram probability. 
Next, the first principal component vector of the sentence embeddings is removed.
The dimensionality of the sentence embeddings created using SIF is equal to that of the pre-trained word embeddings used.
Therefore, both $\vec{w}$ and $\vec{g}(w)$ are in the same $n$-dimensional vector space.
The debiased embedding $E(\vec{w}; \vec{\theta}_{e})$ of $w$ are decoded using a decoder $D_{d}$, parametrised by $\vec{\theta}_{d}$.
The squared $\ell_{2}$ distance between decoded embedding and $\vec{g}(w)$ is computed to define an objective $J_{d}$ given by \eqref{eq:Jd}.
\begin{align}
 \label{eq:Jd}
 J_{d}(w) = \norm{\vec{g}(w) - D_{d}(E(\vec{w}; \vec{\theta}_{e}); \vec{\theta}_{d})}_{2}^{2}
\end{align}

To remove unfair biases from pre-trained word embedding $\vec{w}$ of a word $w$, it is projected into a subspace that is orthogonal to the dictionary definition vector $\vec{g}(w)$.
This projection is denoted by $\phi(\vec{w}, \vec{g}(w)) \in \R^{n}$. 
The debiased word embedding, $E(\vec{w}; \vec{\theta}_{e})$, must be orthogonal to $\phi(\vec{w}, \vec{g}(w))$, and this is formalised as the minimisation of the squared inner-product given in \eqref{eq:Ja}.
\begin{align}
 \label{eq:Ja}
 J_{a}(w) = \left(E(\phi(\vec{w}, \vec{g}(w)); \vec{\theta}_{e})\T E(\vec{w}; \vec{\theta}_{e})\right)^{2}
\end{align}
Note that because $\phi(\vec{w}, \vec{g}(w))$ lives in the space spanned by the original (prior to encoding) vector space, it must be first encoded using $E$ before considering the orthogonality requirement. 

% To derive $\phi(\vec{w}, \vec{s}(w))$, let us assume the $n$-dimensional basis vectors in the $R^{n}$ vector space spanned by the pre-trained word embeddings to be $\vec{b}_{1}, \vec{b}_{2}, \ldots, \vec{b}_{n}$.
% Moreover, without loss of generality,  let the subspace spanned by the subset of the first $k (<n)$ basis vectors $\vec{b}_{1}, \vec{b}_{2}, \ldots, \vec{b}_{k}$ to be $\cB \subseteq \R^{n}$.
% The projection $\vec{v}_{\cB}$ of a vector $\vec{v} \in \R^{n}$ onto $\cB$ can be expressed using the basis vectors as in \eqref{eq:proj}.
% %\par\nobreak
% %{\small
% %\vspace{-5mm}
% \begin{align}
% \label{eq:proj}
% \vec{v}_{\cB} = \sum_{j=1}^{k} (\vec{v}\T\vec{b}_{j}) \vec{b}_{j}
% \end{align}
% %}
% To show that $\vec{v} - \vec{v}_{\cB}$ is orthogonal to $\vec{v}_{\cB}$ for any $\vec{v} \in \cB$, let us express $\vec{v} - \vec{v}_{\cB}$ using the basis vectors as given in \eqref{eq:b1}.
% %\par\nobreak
% %{\small
% %\vspace{-5mm}
% \begin{align}
% \vec{v} - \vec{v}_{\cB} &=  \sum_{i=1}^{n} (\vec{v}\T\vec{b}_{i}) \vec{b}_{i} - \sum_{j=1}^{k} (\vec{v}\T\vec{b}_{j}) \vec{b}_{j} \nonumber \\
%                         &= \sum_{i=k+1}^{n} (\vec{v}\T\vec{b}_{i}) \vec{b}_{i}
% \label{eq:b1}
% \end{align}
% %}
% We see that there are no basis vectors in common between the summations in  \eqref{eq:proj} and \eqref{eq:b1}.
% Therefore, $\vec{v}_{\cB} \T (\vec{v} - \vec{v}_{\cB}) = 0$ for $\forall \vec{v} \in \cB$.
To derive $\phi(\vec{w}, \vec{g}(w))$, \eqref{eq:vb} and \eqref{eq:v_vb} are used.
Considering that $\vec{s}(w)$ defines a direction that does not contain any unfair biases, the vector rejection of $\vec{w}$ on $\vec{g}(w)$ can be computed following this result.\footnote{The rejection of a vector $\vec{b}$ by another vector $\vec{a}$ is defined as $\vec{a} - (\vec{a}\T\vec{b})\vec{a}$.}
Specifically, by subtracting the projection of $\vec{w}$ along the unit vector defining the direction of $\vec{g}(w)$ to compute $\phi$ as in \eqref{eq:phi}.
\begin{align}
\label{eq:phi}
\phi(\vec{w}, \vec{s}(w)) = \vec{w} - \vec{w}\T \vec{g}(w) \frac{\vec{g}(w)}{\norm{\vec{g}(w)}}
\end{align}
The linearly-weighted sum of the above-defined three objective functions is considered as the total objective function as given in \eqref{eq:total}.
\begin{align}
 \label{eq:total}
 J(w) = \alpha J_{c}(w) + \beta J_{d}(w) + \gamma J_{a}(w)
\end{align}
Here, $\alpha, \beta, \gamma \geq 0$ are scalar coefficients satisfying $\alpha + \beta + \gamma = 1$.
Finally, debiased embedding $\vec{d}_{\rm dict}(\vec{w})$ is represented in \eqref{eq:dict}:
\begin{align}
\label{eq:dict}
\vec{d}_{\rm dict}(\vec{w}) := E(\vec{w}; \vec{\theta}_{e})
\end{align}

\end{document}